\documentclass[conference]{IEEEtran}
\IEEEoverridecommandlockouts
\usepackage{graphicx}
\usepackage{amsmath,amssymb} 
\usepackage{color}
\usepackage{subfig}
\usepackage{rotating}
\usepackage{multirow}
\usepackage{array}
\newcolumntype{P}[1]{>{\centering\arraybackslash}p{#1}}
\def\BibTeX{{\rm B\kern-.05em{\sc i\kern-.025em b}\kern-.08em
    T\kern-.1667em\lower.7ex\hbox{E}\kern-.125emX}}
\begin{document}

\title{A Unified Plug-and-Play Framework for Effective Data Denoising and Robust Abstention}
\author{Krishanu Sarker, Xiulong Yang, Yang Li, Saeid Belkasim, Shihao Ji}

\maketitle

\begin{abstract}
Success of Deep Neural Networks (DNNs) highly depends on data quality. Moreover, predictive uncertainty makes high performing DNNs risky for real-world deployment. In this paper, we aim to address these two issues by proposing a unified filtering framework leveraging underlying data density, that can effectively denoise training data as well as avoid predicting uncertain test data points. Our proposed framework leverages underlying data distribution to differentiate between noise and clean data samples without requiring any modification to existing DNN architectures or loss functions. Extensive experiments on multiple image classification datasets and multiple CNN architectures demonstrate that our simple yet effective framework can outperform the state-of-the-art techniques in denoising training data and abstaining uncertain test data. 

\end{abstract}
\section{Introduction}

The immense success of Deep Neural Networks (DNN) in a variety of tasks has caused a revolution in the data-driven learning paradigm~\cite{russakovsky2015imagenet,parkhi2015deep, sarker2018towards,bahdanau2014neural}. However, large amounts of manually annotated data often pose a gridlock constraint towards the success of these deep models. Meta information based automated data collection has been explored as an alternative to manual annotation~\cite{li2017webvision}. However, both types of data acquisition methods are susceptible to error and can introduce noise to the dataset, which results in performance degradation of deep models~\cite{nettleton2010study}. 

DNNs often fall into the issue of erroneous prediction on confusing or noisy in-the-wild samples. This issue reduces the reliability of DNNs even when achieving human-level efficiency. Minimizing predictive uncertainty, hence, is one of the most crucial research problems in order to improve the usability of such deep models in real-world applications, e.g., healthcare systems, autonomous vehicle, secure authentication systems, etc. In this paper, we propose a simple yet effective framework to handle both aforementioned issues in a unified and end-to-end fashion.

For detecting and filtering label noise from data, recent works have utilized a scheme that reduces misclassification loss by incurring penalty while training the model~ \cite{thulasidasan2019combating,geifman2019selectivenet}. SelectiveNet~\cite{geifman2019selectivenet} proposes a specialized rejection model that learns to reject any sample that produces high cross-entropy loss under the constraint of user-specified coverage. The authors show that with different training coverage, inference performance can be improved with corresponding calibrated coverage. DAC~\cite{thulasidasan2019combating} proposed by Sunil Thulasidasan et al. utilizes a similar scheme that abstains hard to learn samples by learning the coverage constraint while training. After introducing artificial noise to data, they show that their method can identify that noise and after filtering them, DNNs can achieve State-Of-The-Art (SOTA) performance. However, these prior works on detecting noisy samples do not evaluate their performance in a unified manner on training data noise detection and abstention of noisy test data. In real-world scenarios, both these issues often coincide. To the best of our knowledge, no unified framework is known to perform well on both tasks in an end-to-end manner.

In this paper, we propose a simple yet effective framework, which can be applied to any SOTA models for both training and test data filtering without any alteration to the model architectures or loss functions. The high-level idea is to model the underlying data distribution in such a way that any sample lying outside known data distribution or sample that is equally distanced from any two or more distributions will be regarded as noise. 

Deep models are proven to learn from dominant features at the beginning of the training process before memorization takes place ~\cite{arpit2017closer,liu2020early}. Hence, deep models, trained with best-practice choices to reduce overfitting, learn robust features even with the presence of noise. We assume that such pre-trained models can learn the underlying data distribution reliably but at a cost of higher error rate. Under this assumption, we utilize the class-specific density of samples in the feature space to identify noisy samples during training and utilize this training data density to identify uncertain samples during inference. With empirical analysis, we observe a strong correlation between the distance of samples from data distribution and the noise associated with them. 

We demonstrate the effectiveness of our proposed framework in both tasks of denoising training data and test data abstention with widely used DNNs on benchmark datasets. Our proposed method outperforms the state-of-the-art method, DAC, on denoising training data. Moreover, we demonstrate the superior performance of our method over SelectiveNet on test data abstention given different coverage calibrations. Through visualization of data samples in feature space, we further justify the effectiveness of our proposed framework.

The contributions of this work are summarized in the following. 
\begin{itemize}
	\item A novel approach to filter noise from both training and test data samples. We propose a density-driven approach for data denoising and abstaining. We introduce modality analysis and adaptive thresholding to differentiate between noise and clean data. 
	\item End-to-end data filtering framework to improve deep models’ reliability in a realistic noisy environment.
	\item Easy to incorporate in any real-world image classification applications as the framework works without any modification to existing SOTA deep models.
	\item Through extensive experimentation and performance analysis, we demonstrate the performance benefit of the proposed framework over existing SOTA methods. 
\end{itemize}

The rest of the paper is organized as follows. In Section \ref{related}, we investigate the current advancement in related research problems. Our methodology will be detailed in Section \ref{method}. Extensive performance analysis is presented in Section \ref{experiment}. Finally, we conclude in Section \ref{conclusion} by mentioning the limitations of this work and future direction we can draw from the experience of this research.

\section{Related Works} \label{related}

\subsection{Selective Prediction}
Classification with a reject option has been explored by researchers to tackle the prediction uncertainty of deep models~\cite{liu2019deep,cordella1995method,de2000reject,bartlett2008classification,grandvalet2009support}. One idea of implementing selective prediction is to define a threshold on posterior probabilities~\cite{cordella1995method,de2000reject}. Various SVMs-style variants have been developed to incorporate reject option with classification tasks~\cite{bartlett2008classification,grandvalet2009support}.

Another more recent trend in this domain is to learn the prediction and the selection parameter jointly \cite{geifman2019selectivenet}. SelectiveNet~\cite{geifman2019selectivenet} proposes a user-defined coverage constraint to learn to abstain samples with high classification loss. By minimizing overall loss, the model learns to abstain from test samples that are difficult to predict. However, in order to set a coverage constraint for the model, users need to have information about the magnitude of noise present in training data, which is infeasible in a real-world scenario. The authors have not demonstrated the effect of noise in training data, which will throw the whole system off the rail, as the model will learn to abstain noisy samples from the training data and potentially misclassify the test samples. One more issue we observed with SelectiveNet is that through the auxiliary head, the system is learning from all examples during the training process even with the presence of noise; this makes the method unsuitable to handle noisy training data. The proposed model requires heavy modifications to existing models and loss functions, which increases overhead. 

\subsection{Label Noise Abstention}
Label noise in training data has received less attention than the selective prediction problem. Yet there are a number of interesting works proposed in the literature to tackle this problem. The authors of~\cite{veit2017learning} have proposed to use two-stream DNN that jointly learns from a large noisy dataset and a small clean dataset. In~\cite{ostyakov2018label}, the authors first train an ensemble of classifiers on data with noisy labels using cross-validation and then the predictions from the ensemble are used as soft labels to train the final classifier. DAC~\cite{thulasidasan2019combating} introduces a more light-weight solution to handle label noise. Unlike SelectiveNet~\cite{geifman2019selectivenet}, DAC proposes to automated learning of noise level while training and use abstention class to determine if a training sample is abstained or not. The authors have empirically shown that adding artifact (smudge) to images results in abstention. However, that might lead to misclassify samples with similar occlusion pattern as noisy, even though the features of the point of interest is still prominent. This might lead to degraded performance with adversarial examples. The authors also have not demonstrated how their proposed model performs when test data were abstained in the presence of label noise.

\subsection{Out-of-Distribution and Adversarial Example Detection}
Out of Distribution detection is another aspect of detecting noise in test data which attracted a lot of attention in recent years~\cite{lee2018simple,liang2017enhancing,shafaei2018less,hendrycks2018deep}. ODIN~\cite{liang2017enhancing} and its variants~\cite{lee2018simple,shafaei2018less} are proven to be very successful in detecting OOD samples. One of the common themes of these methods is the input preprocessing step: adding adversarial noise to test data to increase the difference between in and out-of-distribution data. The authors of~\cite{lee2018simple} have proposed a similar framework to ODIN by adding Gaussian discriminant analysis of samples. They empirically show that Mahalanobis distance can be effective in detecting OOD samples. We adopt the use of distance in detecting noisy data samples, but with key differences from them. For example, we do not employ input preprocessing and we introduce the concept of automated thresholding of distance to differentiate between the noisy and clean examples. However, OOD and adversarial sample detection are out of the scope of the current work. We will address these issues in the future.

\section{Methodology} \label{method}
Given a DNN architecture, we propose a simple yet effective framework for detecting noise in data. First, we present our intuition behind the core of the proposed framework. Then we define the algorithmic steps of the framework in details. 

\begin{figure*}[h]
\centering
    \subfloat[]{{\includegraphics[width=4cm]{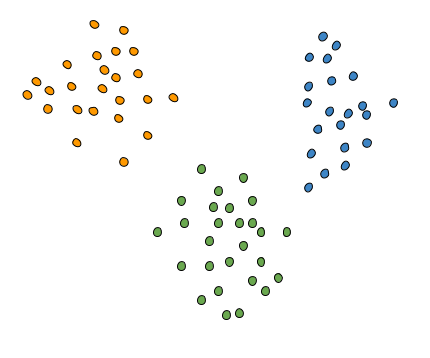} }}%
    \qquad
    \subfloat[]{\includegraphics[width=4cm]{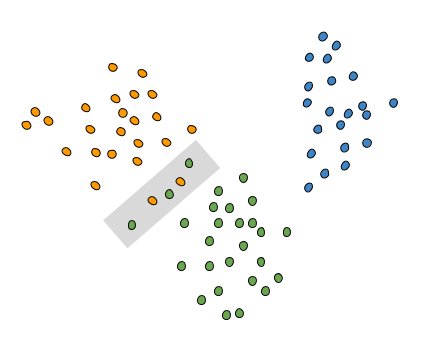} }
    \subfloat[]{\includegraphics[width=4cm]{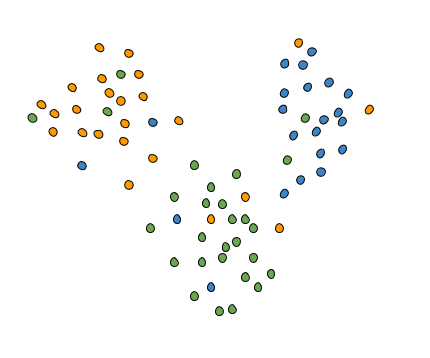} }%
\caption{Dummy data distributions simulating different learning scenarios: (a) in an ideal scenario, (b) in presence of confusing samples that lie in the border of two distributions (highlighted in gray), and (c) in presence of label noise, where a fraction of samples are mislabeled.}
\label{img:dummy_formulation}
\vspace{-0.5cm}
\end{figure*}

We present an analogy of human social behavior to explain our intuition of noise in data. A group of people who share common interests are more likely to spend more time together or do more interaction with each other. Conversely, a group of people who do not have common interests are less likely to be together. Similarly, data samples that share dominant features are more likely to belong to the same cluster or class and data samples that project contrasting features are less likely to belong to the same distribution. Based on this hypothesis, we construct a framework to differentiate noise from data. 

\textbf{Hypothesis:} \textit{Samples that are away from distribution are potentially noisy or mislabeled.}

DNN learns high-level features from data samples during training. In the case of supervised learning, these features follow class constraints provided by labels. These features are high dimensional vectors that represent each data sample. To simplify, let us consider a dummy dataset where each sample only consists of two features. In an ideal world, samples of each class would be clearly separated and all samples of a class would be clustered together. We define these two ideal situations as inter-class diversity and intra-class affinity respectively. However, in real-world data, there exist inter-class affinity and intra-class diversity often due to errors in labeling or noise in data samples (Fig. ~\ref{img:dummy_formulation}).

Let us consider a training set consisting of input-target pairs, \(D = {(x_i, y_i)}^N _{i=1}\), where \(x_i \in \mathbb{R}^n\) belongs to one of the $k \in L=\{l_1,l_2,\cdots, l_k\}$ classes. Note that in this paper we will state ``class" and ``cluster" interchangeably, where both of them are semantically similar. A DNN classifier consists of a feature extractor and a classifier. Feature extractor is a parameterized function \(f_\theta : \mathbb{R}^n \rightarrow \mathbb{R}^\delta\) that learns to map \(n\) dimensional observed data \(x_i\) to feature space \(v_{x_i}\) of \(\delta\) dimensions under the \(l_j \in L\) class constraint. Classifier is a simple mapping function, \(f_c: f_\theta(x_i) \rightarrow y_i\), which can be a softmax classifier. 

Typically, parameter \(\theta\) is optimized using the off-the-shelf cross-entropy loss. In an ideal scenario, \(f_\theta\) thus learns to cluster semantically-similar inputs \(x_i\) to \(k\) different clusters corresponding to $k$ classes. For any given new sample \(s\), \(f_\theta\) maps it to a feature vector \(v_{s}\), which ideally should lie under any of \(k\) distributions observed during training. However, in real-world data, there might be \(p \ge k\) clusters formed by observed data samples of \(k\) classes. Our goal here is to define each of these \(k\) data distributions robustly, such that even with presence of noise, definitions of these distributions hold. We propose to utilize density based clustering to identify which of these \(p\) clusters actually represent \(k\) classes and then we calculate centroids to represent these \(k\) clusters.

\subsubsection{Density-based Clustering} To identify core classes from \(p\) clusters, we propose to use DBSCAN clustering~\cite{ester1996density}, on feature space. Let us assume there are \(N_j\) samples that are bounded by the same class constraint \(l_j\). Feature vectors \(v_{x^j_i}\) extracted by \(f_\theta\) are utilized with \(DBSCAN\) algorithm to identify hidden clusters within class \(l_j\). Any sample that is not affine to \(MinPts\) number of density-reachable samples are treated as noise, and samples that are affine to at least \(MinPts\) number of samples form a cluster. Though, there can be multiple clusters detected by \(DBSCAN\) within a given class constraint, we define the cluster with the highest samples as the core cluster. The rational behind is that only the most populous cluster can be representative enough of a particular class. Other clusters with less samples may potentially be label noise occurred during data acquisition. However, in scenarios where there is no label noise presents, we expect to see a single cluster from the \(DBSCAN\) algorithm. Identifying the most representative density distribution is crucial for our proposed framework, as we will define this cluster as the reference point. 

\subsubsection{Calculation of Centroid} \label{centroid_calc} 
We calculate the centroid of each class constrained core cluster yielded from \(DBSCAN\) by calculating the median of feature vectors \(v_x\) extracted from trained DNN for sample \(x\). Let us assume there are \(m\) samples in class \(j\), and \(DBSCAN\) returns a core cluster with \(m_{core}\) samples, where \(m_{core} \le m\). Then the centroid is defined as
\begin{equation} \label{eq:centroid}
 c_j = \text{median}([{v_{x^j_i}}]^{m_{core}}_{i=0}), \;\;\;\;\forall x^j_i \in l_j.  
\end{equation}
We collectively denote all $k$ identified centroids as $C=\{c_1, c_2,\cdots,c_k\}$.

We take the approach of refining the data in multiple stages. Broadly, this can be divided into two stages. Firstly, we conduct denoising training data by utilizing a pretrained model and then we dive into abstaining from inferring noisy or confusing test data in inference time. 

\subsection{Denoising Training Data} \label{denoise} 
In the first stage, we calculate distance between data samples and observed distributions and filter based on the derived distances. The first stage can be further granulated into five steps.

Step 1: We first train a DNN model with given training data (noisy or clean) with regularization. As demonstrated in~\cite{arpit2017closer}, deep models learn from dominant features at the beginning of training. We confirm that claim empirically by training models with smaller number of epochs before memorizing starts. We also follow ``best practice" to reduce overfitting.

Step 2: We employ \(DBSCAN\) on \(m\) samples belong to each class to identify the core cluster with \(m_{core}\) samples. Then we calculate centroids for each class using Eq.~\ref{eq:centroid}.

Step 3: We calculate the distance \(d^j_{x_i}\) between the feature vectors \(v_{x^j_i}\) with label \(l_j\) and the corresponding centroid \(c_j \in C\). In this step, we consider all \(m\) samples that belong to class \(j\). We choose to use euclidean distance as our distance measure.  
\begin{equation} \label{eq:train_dist}
d^j_{x_i} = \text{euclid}(v_{x^j_i}, c_j)
\end{equation}

Step 4: We propose a methodology to denoise any outliers by multimodality analysis, which will be discussed in details in Section \ref{modality}. 

Step 5: We train the model from scratch with denoised data we derived from the previous step. 

\subsection{Abstain from Inferring on Test Data} \label{abstain}
Second stage of our proposed framework takes place during the inference. At this stage, we already have a trained model on \emph{cleaned} data. The second stage can be further divided into four steps. 

Step 1: We calculate distance between all test samples and training data distributions under constraint of \(k\) classes. For \(s \in S\), where \(S\) is the set of in-the-wild test samples, we calculate distance \(d_{s}\) between \(s\) and all \(c_j \in C\) we derived in the previous stage:
\begin{equation} \label{eq:test_dist}
  d^j_{s} = \text{euclid}(v_{s}, c_j).  
\end{equation}
Here each sample \(s\) will have \(k\) distance values each corresponding to the distance from \(k\) classes. Note that the difference between Eq.~\ref{eq:train_dist} and Eq.~\ref{eq:test_dist}: we do not have the class label information for test sample \(s\), whereas we know the ground-truth label for training sample \(x_i\).

Step 2: We propose to invoke our first filtering criterion on test data based on the distance we calculated in the previous step. It is expected that trained models can make better predictions when test data follows the similar distribution as the training data. However, for a state-of-the-art DNN, it is not possible to differentiate between samples that do or do not belong to the same distribution as it has observed during the training process. Hence, we calculate the maximum distance observed from respective centroids in training data to get the sense of data distribution. We utilize this maximum distance as a threshold \(\tau\) for in-the-wild test samples so that the model can identify out-of-distribution samples. Specifically, we calculate $\tau$ as follows
\[\tau_j = \max([{d^j_{x_i}}]^{N_j}_{i=0}),\;\;\;\;\forall x_i^j\in l_j\] where \(d^j_{x_i}\) is the distance between centroid \(c_j\) and train sample \(x_i\) that belong to class \(l_j\).

Step 3: In the first phase of two layered filtering, we abstain test samples based on the threshold we calculate from training data. We first get the minimum of distances between each test sample \(s\) and all clusters \(c_j \in C\). With this step, we abstain from classifying out-of-distribution samples. 
\begin{align}
    d^{min}_{s} &= \min([{d^j_{s}}]^k_{j=0})\\
     c^{min} &= \arg\min([{d^j_{s}}]^k_{j=0})
\end{align}
And we abstain samples if the following condition satisfies:
\begin{equation}
    d^{min}_{s} > \tau_{c^{min}}
\end{equation}

Step 4: Having out-of-distribution samples abstained, we here focus on the noisy or confusing samples. Samples that are similarly distanced from two or more class-constrained data distributions, are deemed as confusing samples. We only consider those distributions that are closest from the sample since samples belong to the distribution that they are closest to. We abstain sample \(s\) if the following condition holds: 
\[|d^{a}_{s} - d^{b}_{s}| > \eta, \]
where \(a\) and \(b\) are the two nearest clusters from sample \(s\). \(d^{a}_{s}\) and \(d^{b}_{s}\) are the distances between sample \(s\) and centroids \(c_{a}\) and \(c_{b}\), respectively, and \(\eta\) is a tolerance parameter that we set empirically.

\subsection{Modality Detection and Thresholding} \label{modality}

\begin{figure*}[h]
\centering
    \subfloat[Distribution without noise]{{\includegraphics[width=5.7cm]{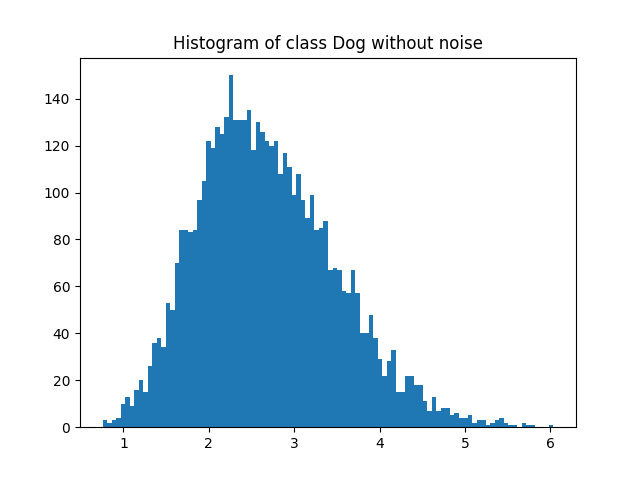} } }%
    \qquad
    \subfloat[Distribution with 20\% noise]{{\includegraphics[width=5.7cm]{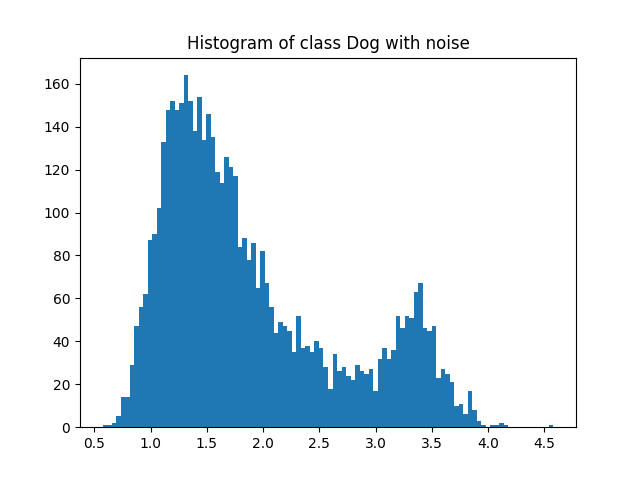} } }%
    \qquad
    \subfloat[PDF and Otsu's threshold on distribution]{{\includegraphics[width=5.7cm]{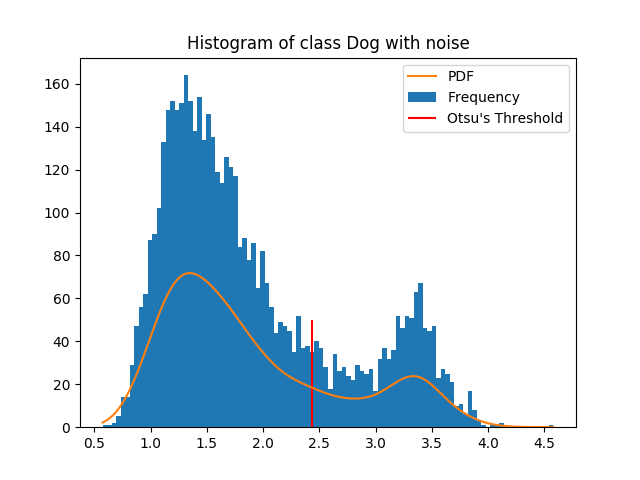} } }%
    \qquad
    \subfloat[Distribution of 20\% noise]{{\includegraphics[width=5.7cm]{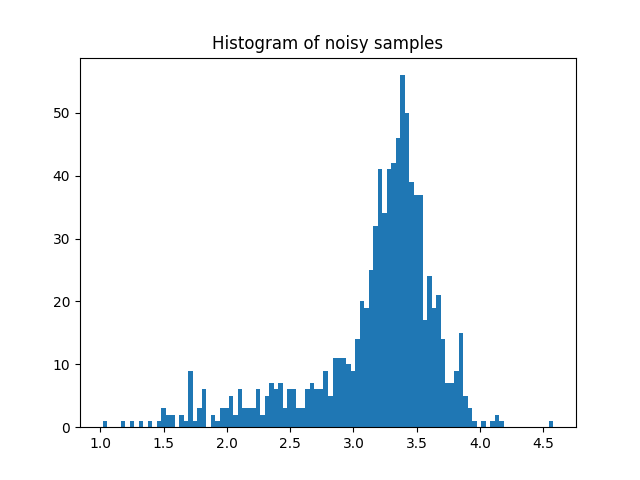} } }%
    \qquad
\caption{Histograms of the distances between samples from CIFAR10 dataset and cluster centroids. (a) Histogram of the distances between samples of ``Dog" and centoid without presence of noise. We can observe unimodal distribution. (b) Histogram of the distances between samples of ``Dog" and centroid with presence of 20\% random label noise. A bimodal distribution is observed. (c) Histogram and PDF (estimated by KDE) of the distance distribution. Red vertical line represents the Otsu's adaptive threshold for the given distribution. (d) Histogram of the distances between noisy samples and centroid. We can observe that Otsu's method can reliably identify a cutoff that differentiates noise from clean data samples.}
\label{img:euclid1}
\vspace{-0.4cm}
\end{figure*}

We hypothesize that DNN features are closely clustered when samples share similar features, and they become scattered when there are less correlations between features. When noise is present in any classes of data, varying correlations between samples are observed. For example, as depicted in Fig.~\ref{img:euclid1}(b), multiple modalities in distance distribution of noisy samples from the same class are observed. We also observe from Fig.~ \ref{img:euclid1}(a) single modality in distance distribution when data from the same class is free of noise, which supports the above hypothesis. Modality in distance distribution plays a key role in detecting noisy samples during training. 

We utilize Kernel Density Estimation (KDE) to perform the modality test on distance distribution. Distance \(d\) here can be considered as a univariate sample that is drawn from some distribution with unknown density that we would like to model. With KDE, the Probability Density Function (PDF) of \(d\) can be approximated as
\begin{equation}
    \text{PDF}(d)\approx \frac{1}{nh}\sum_{i=1}^{n}K\left ( \frac{d-d_i}{h} \right ),
\end{equation}
where \(K\) is the kernel function, e.g. the \text{Gaussian kernel}, and \(h\) is a smoothing parameter that we empirically set to 0.3 in order to avoid detecting false peaks.

We identify the number of peaks by calculating the gradient of the KDE curve. If we detect a single peak, we can define the distribution as unimodal, otherwise multimodal. Interestingly, in all our experiments we have observed that in presence of random label noise, distance distributions always follow bimodality. Hence, here we focus on bimodal distributions. But our proposed method can easily be generalized to multimodal distributions. 

In case of a bimodal distance distribution, we propose to re-purpose Otsu's thresholding~\cite{otsu1979threshold} to detect cut-off threshold in order to detect noisy samples from training data. In image processing, Otsu’s method is widely used to perform automated image binarization. The algorithm returns a single intensity threshold to separate image pixels into two classes: foreground and background. The algorithm exhaustively searches for a threshold \(t\) that maximizes the inter-class variance \(\sigma^2_B\) of the two classes, which is defined as 
\begin{equation}
    \sigma^2_B(t) = \omega_{0}(t) \omega_{1}(t)(\mu_1(t) - \mu_0(t))^2
\end{equation}
where \(\omega_0\) and \(\omega_1\) are the probabilities of the two classes separated by \(t\) and \(\mu_0\) and \(\mu_1\) are the means of two classes. We repurpose this algorithm to detect the cut-off threshold of bimodal distance distribution. We define two classes from Otsu’s algorithm as clean and noisy data distributions. Fig.~\ref{img:euclid1}(c) shows the detection of bimodality by detecting peaks in distance distribution of noisy data; it also shows modality testing and Otsu’s thresholding in practice. As we can see, Otsu’s thresholding can effectively identify cut-off value to differentiate between noisy data and clean data. Compared to Fig.~\ref{img:euclid1}(d), which illustrates the ground truth distance distribution of randomized samples, we can observe that a very small number of noisy data samples fall below the Otsu's threshold, hence not excluded from training set. We deem this as expected since with 20\% label noise introduced randomly, the probability of samples not being randomized within this 20\% for a particular class is $\frac{1}{k}$ given that we are randomizing each of the \(k\) classes uniformly.

\section{Experimental Analysis} \label{experiment}
In this section we demonstrate the performance of the proposed framework using various CNN architectures, e.g. VGGNet~\cite{simonyan2014very} and ResNet~\cite{he2016deep} on multiple image classification benchmarks: CIFAR10~\cite{krizhevsky2009learning}, SVHN ~\cite{netzer2011reading}, and Fashion-MNIST~\cite{xiao2017fashion}. We compare our method with the state-of-the-art algorithms SelectiveNet~\cite{geifman2019selectivenet} and DAC~\cite{thulasidasan2019combating}. To ensure a fair comparison, our experiments closely follow those of the competing methods. We plan to open source our code to facilitate the research in this area. 

\subsection{Detecting label noise}
We aim at a problem of label noise that might occur on some fraction of data. Here, we assume that a fraction of labels have been corrupted by random assignment. Our proposed framework identifies the mislabeled samples and removes them as noisy samples from training set. To identify the corrupted samples, we first train an off-the-shelf DNN with best practice regularization to avoid overfitting using a validation set, which we assume to be clean. Our proposed framework utilizes the features extracted by the trained DNN to differentiate between noisy and clean training examples. We present our results by retraining the same DNN from scratch with cleaned training set. 

We first compare our proposed framework with DAC ~\cite{thulasidasan2019combating}, a state-of-the-art method that introduces an additional abstention class to learn to abstain noisy samples during training. We also present the performance of the bare baseline model, which is the same DNN utilised in both DAC and our proposed method. To ensure fairness, we report our results using similar setup as~\cite{thulasidasan2019combating} and we use the numbers reported in their paper~\cite{thulasidasan2019combating}.

We conduct experiments on CIFAR10~\cite{krizhevsky2009learning} and Fashion-MNIST~\cite{xiao2017fashion} with varying amount of arbitrary label noise. In our proposed approach, we use same CNN architectures for pretraining and retraining phases. We use the same hyperparameters, e.g., initial learning rate, learning rate decay and optimizer, as in DAC and baseline model for the retraining phase. We utilize ResNet18 and ResNet34~\cite{he2016deep} without modifications as our DNN architecture for experiments presented in this section. For the \(DBSCAN\) algorithm, we empirically set \(MinPts = 300\) and \(eps = 0.8\) for our experiments. We randomly choose a seed value (seed = 1) in our experiments to ensure reproduciblity.

\begin{table} 
\centering 
\begin{tabular}{P{.9cm}P{.8cm}P{1.5cm}P{1.5cm}P{1.4cm}} 
\hline
\cline{1-5}
\multicolumn{1}{c}{}&\multicolumn{1}{c}{} &\multicolumn{3}{c}{Models}              \\ 
\cline{3-5}
Dataset &\multicolumn{1}{c}{Noise Label} & Baseline & DAC     & Ours\\ 
\cline{1-5}
\hline
\multirow{7}{*}{\begin{tabular}[c]{@{}c@{}}CIFAR10\\(ResNet34)\end{tabular}} & 20\%                  & 88.64\%  & \begin{tabular}[c]{@{}c@{}} 92.91\%\\(0.24/0.01)\end{tabular} & \begin{tabular}[c]{@{}c@{}} \textbf{93.03\%}\\(0.25/0.03)\end{tabular} \\ 

& 40\%                  & 85.95\%  & \begin{tabular}[c]{@{}c@{}}90.71\%\\(0.41/0.03)\end{tabular} & \begin{tabular}[c]{@{}c@{}} \textbf{90.88\%}\\(0.41/0.03)\end{tabular} \\ 

& 60\%                  & 80.92\%  & \begin{tabular}[c]{@{}c@{}}\textbf{86.30\%}\\(0.56/0.07)\end{tabular} & \begin{tabular}[c]{@{}c@{}} 86.28\%\\(0.56/0.05)\end{tabular} \\ 

& 80\%                  & 67.17\%  &  \begin{tabular}[c]{@{}c@{}}\textbf{74.84\%}\\(0.75/0.16)\end{tabular} & \begin{tabular}[c]{@{}c@{}} 69.7\%\\(0.64/0.16)\end{tabular} \\
\cline{1-5}
\hline
\multirow{7}{*}{\begin{tabular}[c]{@{}c@{}} Fashion-\\MNIST\\(ResNet18)\end{tabular}} & 20\%                  & 93.92\%  & \begin{tabular}[c]{@{}c@{}} 94.76\%\\(0.25/0.01)\end{tabular} & \begin{tabular}[c]{@{}c@{}} \textbf{94.95\%}\\(0.21/0.01)\end{tabular} \\ 

                          &40\%                  & 93.09\%  & \begin{tabular}[c]{@{}c@{}}94.09\%\\(0.48/0.01)\end{tabular} & \begin{tabular}[c]{@{}c@{}} \textbf{94.20\%}\\(0.38/0.02)\end{tabular} \\ 

                 &60\%                  & 91.83\%  & \begin{tabular}[c]{@{}c@{}}92.97\%\\(0.66/0.03)\end{tabular} & \begin{tabular}[c]{@{}c@{}} \textbf{93.05\%}\\(0.58/0.01)\end{tabular} \\ 

                         &80\%                  & 88.61\%  &  \begin{tabular}[c]{@{}c@{}}\textbf{90.79\%}\\(0.88/0.04)\end{tabular} & \begin{tabular}[c]{@{}c@{}} 89.77\%\\(0.72/0.03)\end{tabular} \\
\cline{1-5}
\hline
\end{tabular} 
\caption{Comparative results with varying percentages of noise labels. We compare our proposed framework with a baseline model (off-the-shelf DNN) and DAC~\cite{thulasidasan2019combating}. The numbers in parenthesis indicate the fraction of data removed and the remaining label noise.}
\label{tbl:labelnoise} 
\vspace{-.55cm}
\end{table}

Table~\ref{tbl:labelnoise} presents the comparative results of this experiment. Our proposed framework achieves improved accuracy in most of our experiments as compared to the state-of-the-art DAC~\cite{thulasidasan2019combating}. Our framework could identify noisy data points reliably, even outperforming specialized learning model DAC when percentage of noise label is lower than 60\%. When percentage of noise label is 80\%, we observe that our framework does not perform as well as DAC. This is because with highly corrupted data our method's performance degrades as DNN struggles to learn class specific features, which results in scattered feature distribution of training data. In order to promote simplicity, our framework does not use the feedback loop from the noise to the model, whereas DAC~\cite{thulasidasan2019combating} models the noise explicitly while training, which helps them to learn more from the clean samples than from the noisy ones, yet DAC still suffers from the issue of memorization~\cite{thulasidasan2019combating}. Nevertheless, we argue that presence of very high amount of noise (e.g., 80\%) in dataset is not very realistic in real world as the label quality in this case is close to be random (e.g., random guess accuracy on 10 classes is already 10\%), hence investing heavily to improve performance in this scenario is impractical. Despite that, our framework achieves improved results when percentage of noise label is lower than 60\% 
even though DAC requires specialized loss function to learn the pattern of noise while training, whereas our proposed framework employs simple yet effective filtering approach on feature space extracted by DNNs.

\begin{table}[t]
\centering
\begin{tabular}{P{.75cm}P{.7cm}P{1.8cm}P{1.8cm}P{1.7cm}} 
\cline{1-5}
\hline
 & & \multicolumn{3}{c}{Models}         \\ 
\cline{3-5}
Dataset & Coverage & \begin{tabular}[c]{@{}c@{}}SelectiveNet\\(varying coverage)\end{tabular}      & \begin{tabular}[c]{@{}c@{}}SelectiveNet\\(100\% coverage)\end{tabular}  & Ours       \\ 
\cline{1-5}
\cline{1-5}
\multirow{7}{*}{\begin{tabular}[c]{@{}c@{}} CIFAR10 \\(VGG16)\end{tabular}} & 100\%        &  93.21\%       & 93.21\%             & 93.21\%           \\ 

& 95\%         &  95.40\%         & 95.44\%              & \textbf{95.64\%}  \\ 

& 90\%         &  97.27\%         & 97.16\%             & \textbf{97.41\%}  \\ 

& 85\%         &  \textbf{98.40\%}         & 98.19\%              & \textbf{98.40\%}  \\ 

& 80\%         & \textbf{99.03\%} & 98.69\%     & 98.97\%           \\ 

& 75\%         & \textbf{99.31\%} & 98.83\%    & 99.24\%           \\ 

& 70\%         & \textbf{99.40\%}         & 98.94\%              & \textbf{99.40\%}   \\
\hline
\multirow{5}{*}{\begin{tabular}[c]{@{}c@{}} SVHN \\(VGG16)\end{tabular}} & 100\%        &  96.22\%       & 96.22\%             & 96.22\%           \\ 

& 95\%         &  \textbf{98.20\%}                       & 97.80\%              & 97.88\%  \\ 

& 90\%         &  98.97\%                                & 98.74\%              & \textbf{99.07\%}  \\ 

& 85\%         &  99.25\%                                & 98.99\%              & \textbf{99.40\%}  \\ 

& 80\%         &  99.41\%                                & 99.10\%              & \textbf{99.49\%}           \\ 

\hline
\end{tabular}
\caption{Comparative results on CIFAR10 with varying calibrated coverages of our proposed framework and SelectiveNet~\cite{geifman2019selectivenet}. We highlight the best performances with boldface. Note that numbers presented here on SelectiveNet (column one) obtained by training their model with varying user specified coverages and then calibrated with corresponding coverage value. Results obtained from our proposed framework required training the DNN only once. We performed calibration on that trained DNN only. Hence, for fair comparison, we include SelectiveNet (column two) that presents results when trained their model with 100\% coverage, similar to ours.}
\label{tbl:selective} 
\vspace{-.55cm}
\end{table}

\subsection{Abstaining test samples}
We now consider the predictive uncertainty problem during inference. For these particular experiments, we assume training data is free of noise, but in-the-wild test samples may be noisy or confusing. We aim to abstain such samples using our proposed framework to reduce predictive uncertainty. We first train an off-the-shelf DNN with given dataset and in the post training phase we employ our algorithm to filter out samples that are deemed confusing or out-of-distribution. 

To demonstrate the advantages of our proposed framework, we compare its performance with state-of-the-art SelectiveNet ~\cite{geifman2019selectivenet}, and report the results in Table~\ref{tbl:selective}. SelectiveNet~\cite{geifman2019selectivenet} proposes a specialized rejection model that learns to reject any sample that produces high cross-entropy loss under the constraint of user-specified coverage. We use similar parameter settings reported in the paper~\cite{geifman2019selectivenet} for a fair comparison. Note that reported numbers for ``SelectiveNet (varying coverage)" are obtained by training with target coverage value and inferred on the same calibrated coverage as described in~ \cite{geifman2019selectivenet}, whereas we report our performance by training DNN once and use varying calibrated coverage by tuning tolerance parameter \(\eta\) accordingly only during inference. To make a fair comparison, we also train SelectiveNet with 100\% coverage only, similar to ours, and then use varying calibrated coverage to obtain results for ``SelectiveNet (100\% coverage)". Our proposed framework achieves reported results with greatly reduced complexity (both time and resource) compared to SelectiveNet. 

\begin{figure*}[t]
\centering
    
    \subfloat[Training-set features without label noise]{\includegraphics[width=5cm]{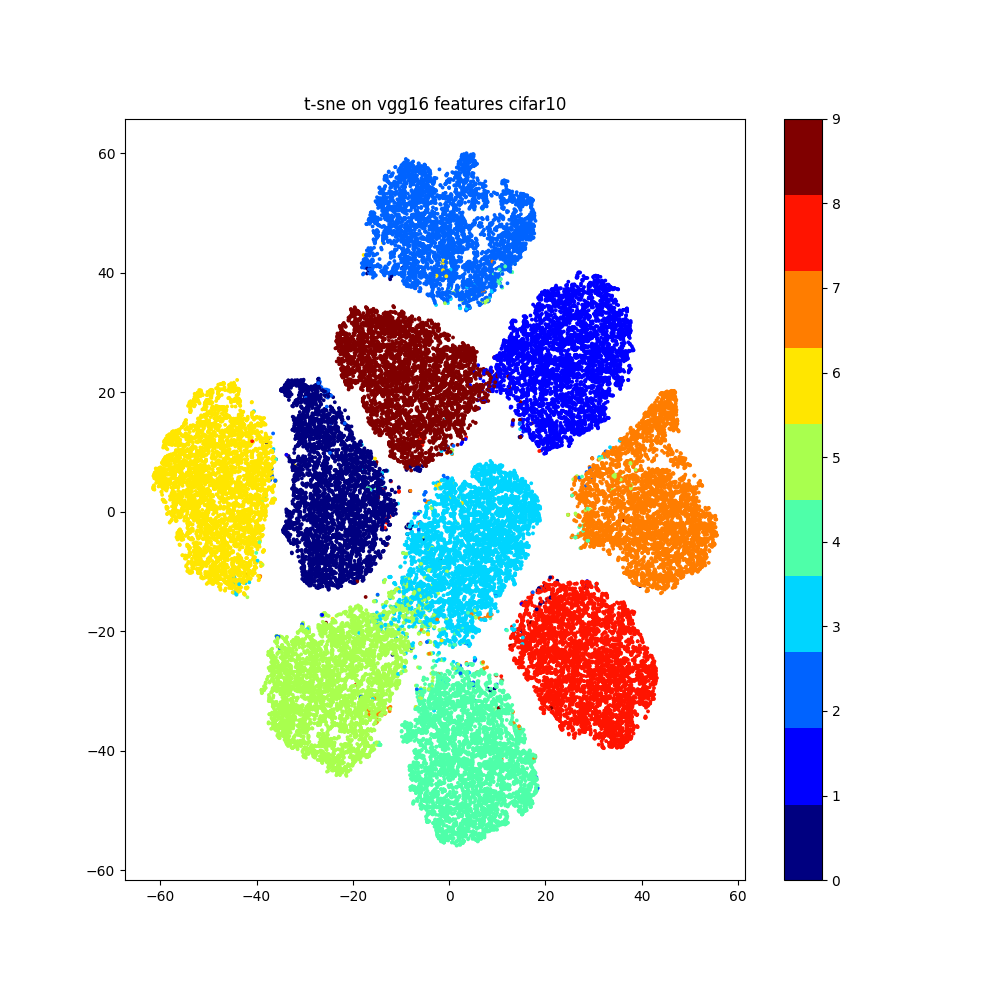} }%
    \qquad
    \subfloat[Training-set features with 20\% label noise]{\includegraphics[width=5cm]{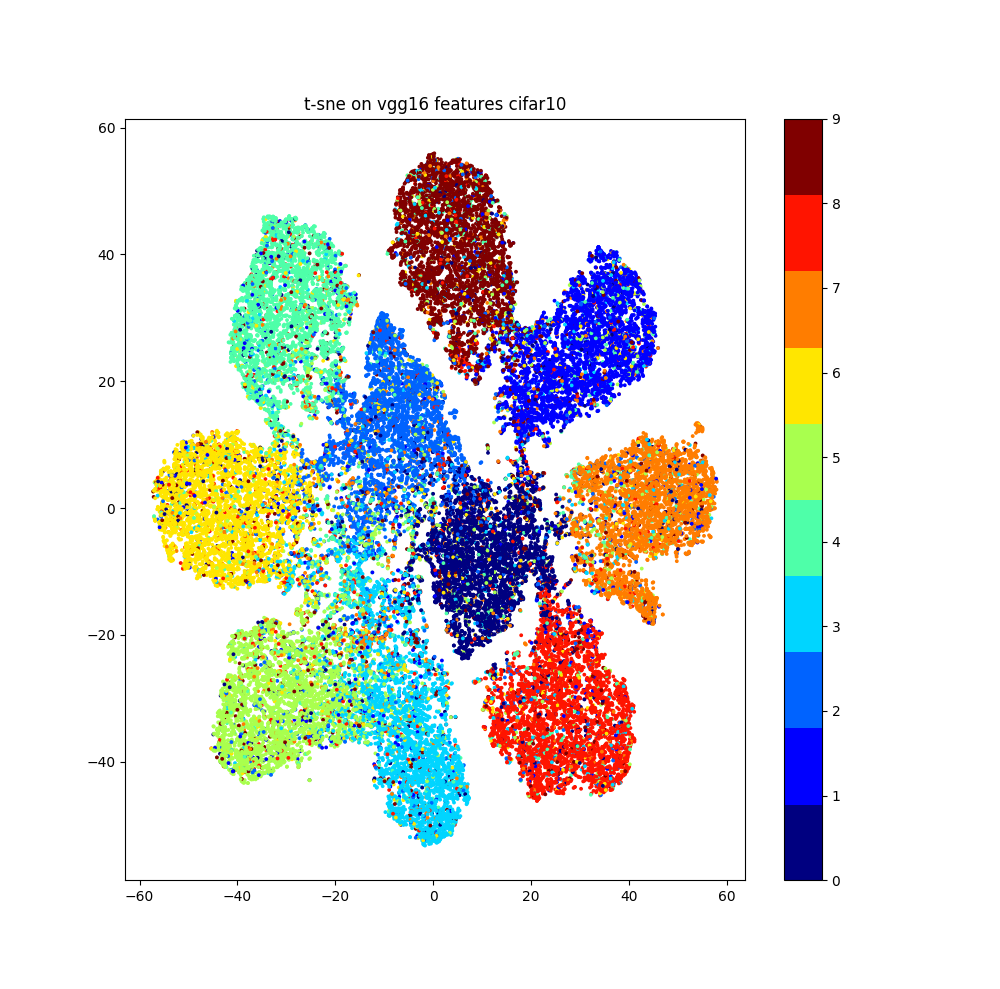} }%
    \qquad
    \subfloat[Training-set features after employing our framework  ]{\includegraphics[width=5cm]{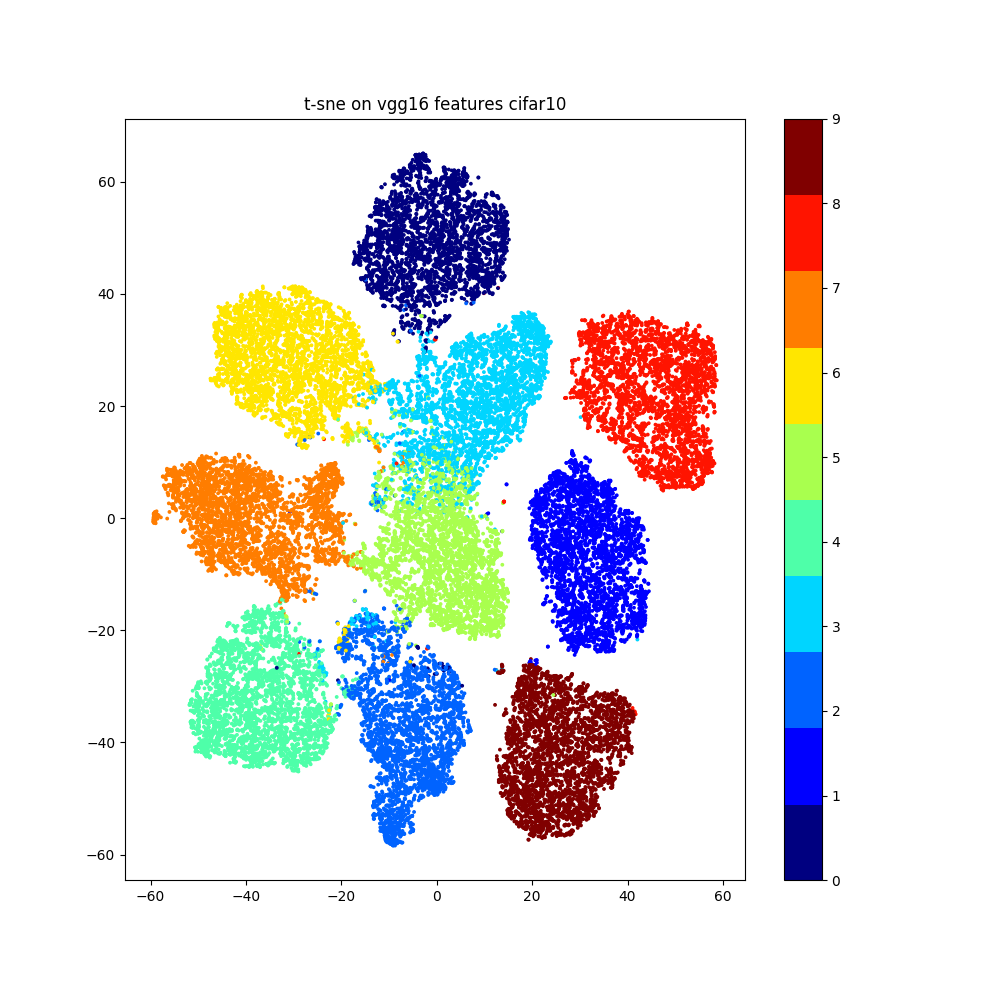} }%
    \qquad
    \subfloat[Test-set features from DNN trained with noisy data]{\includegraphics[width=5cm]{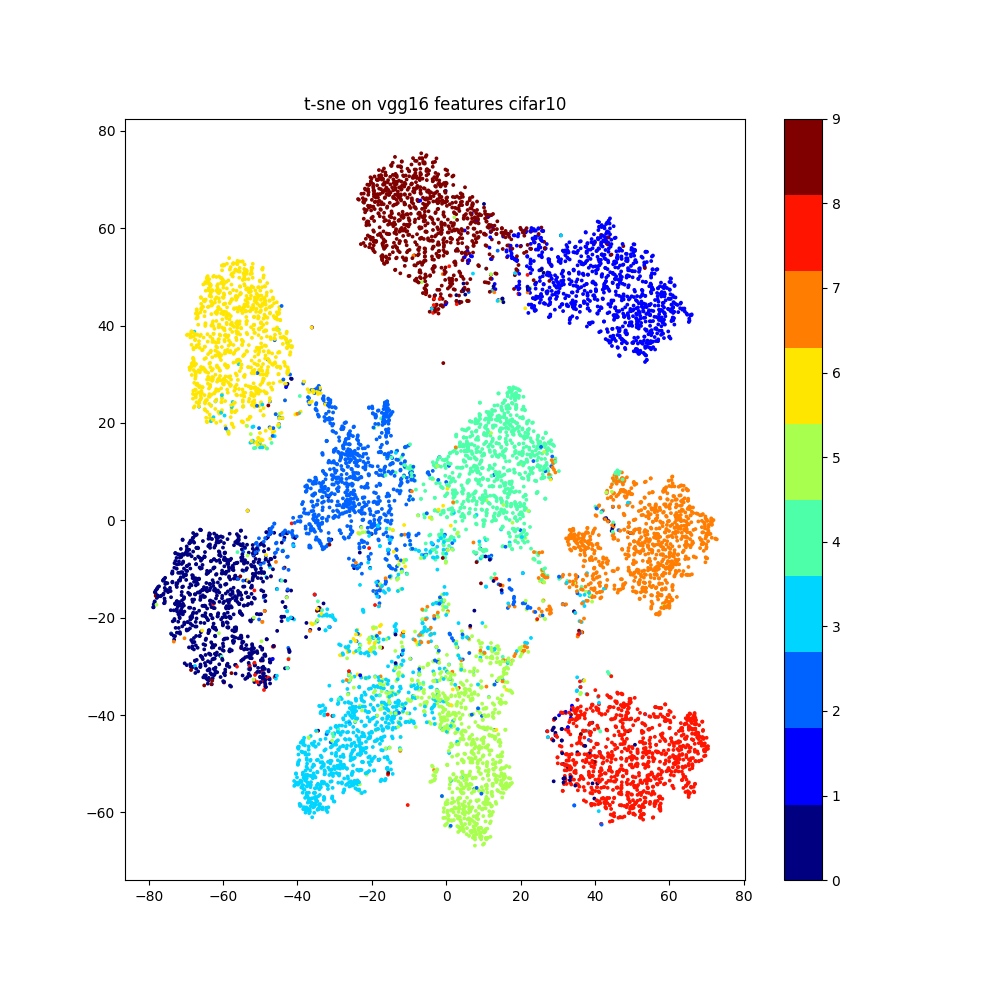} }%
    \qquad
    \subfloat[Test-set features from DNN trained with denoised data]{\includegraphics[width=5cm]{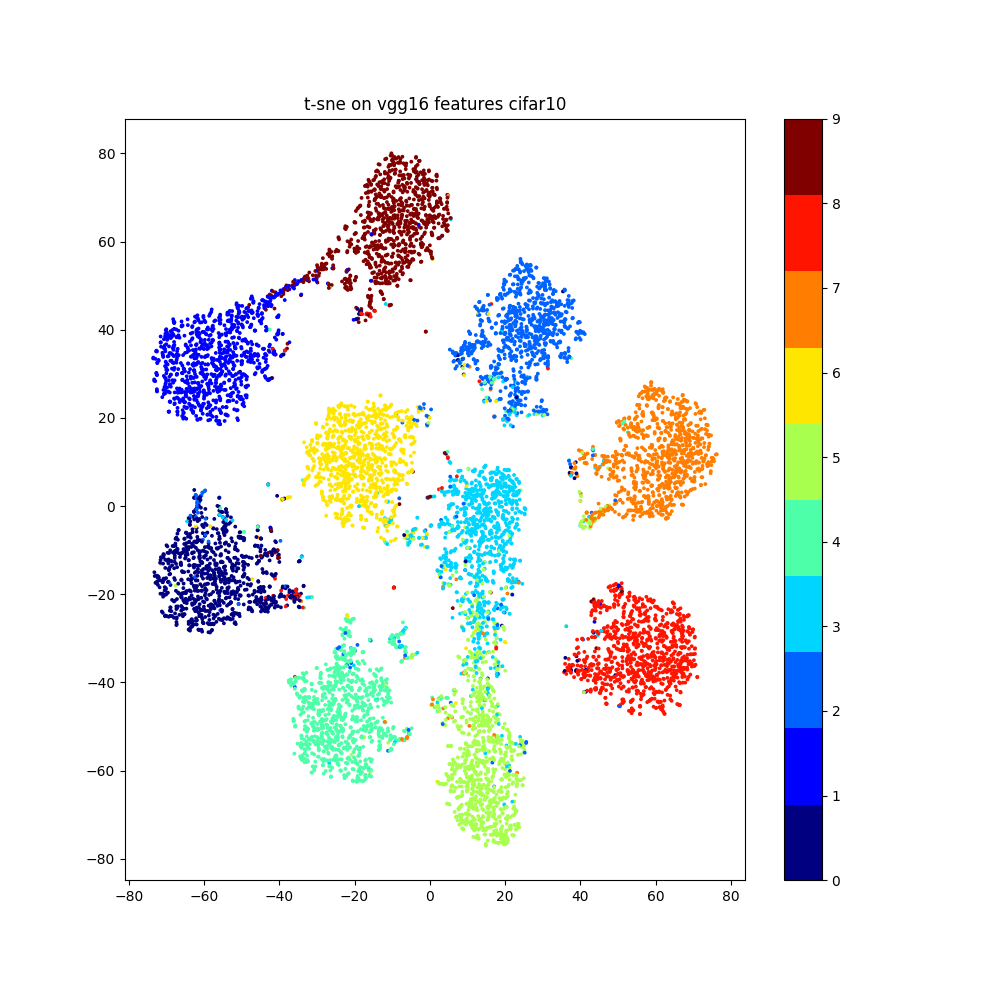} }%
    \qquad
    \subfloat[Test-set features after abstaining from DNN trained with denoised data]{\includegraphics[width=5cm]{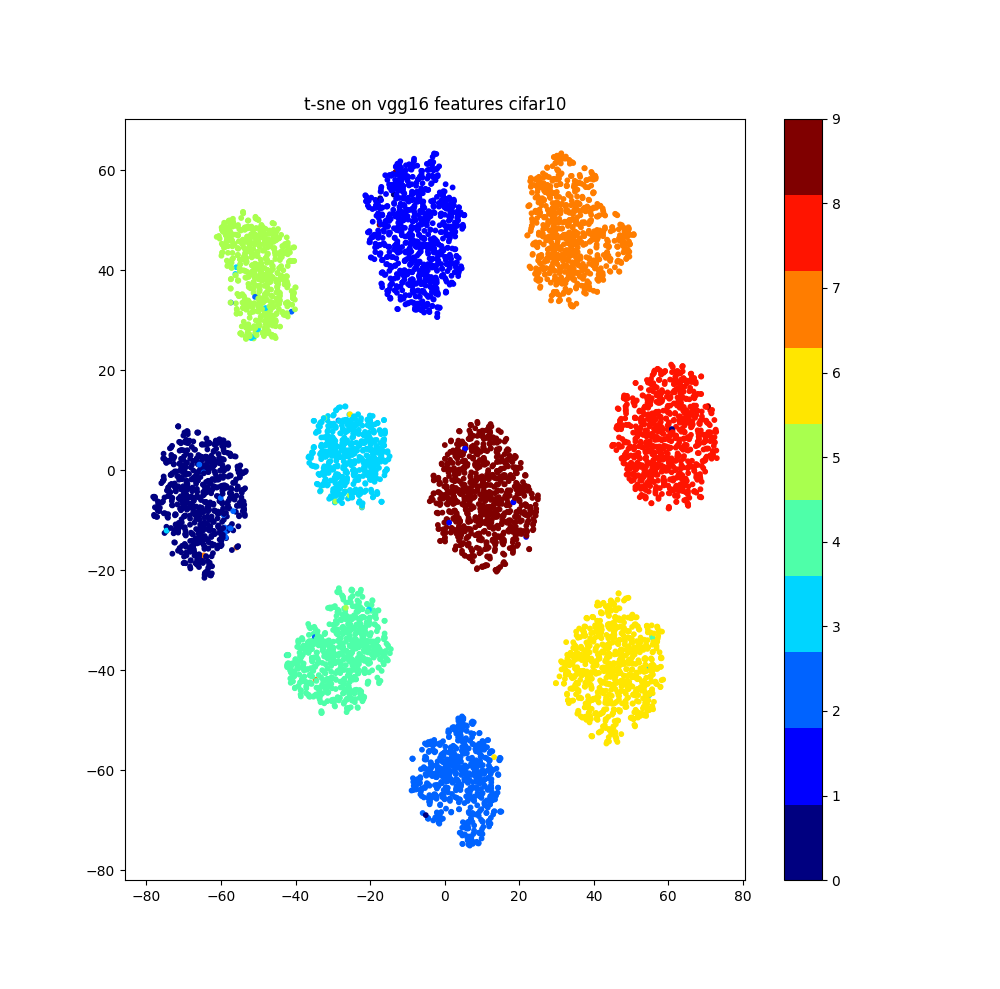} }%
    \qquad
\caption {t-SNE visualization of CIFAR10 training and test sets in feature space.}
\label{img:tsne}
\vspace{-.5cm}
\end{figure*}

We conduct experiments on CIFAR10~ \cite{krizhevsky2009learning} and SVHN~ \cite{netzer2011reading} with off-the-shelf VGG16 \cite{simonyan2014very} architecture. Performance analysis is presented in Table~\ref{tbl:selective}. We observe that our proposed framework can outperform or achieve very similar performance compared to SelectiveNet~ \cite{geifman2019selectivenet} for both datasets. Moreover, our framework demonstrates a very clear advantage when compared with results from ``SelectiveNet (100\% coverage)". We found this observation very intriguing as our proposed method only takes advantage of feature learning capability of DNN coupled with intuitive filtering techniques. This would mean that using specialised loss functions to abstain samples has very small impact on the performance, and DNNs are robust enough to learn distinctive features but lack the ability to reject noisy or confusing samples. Overall, not only being more efficient (i.e., training once), our proposed framework also achieves better accuracies in most of the coverage levels, demonstrating the superiority of our method.
%
%
%
%

\subsection{Visualizing Effectiveness of Proposed Framework}
In order to demonstrate the effectiveness of our proposed framework in detecting noise in both training data as well as in-the-wild test data, we visualize the feature spaces of the trained model ResNet34~\cite{he2016deep} on CIFAR10~\cite{krizhevsky2009learning} using T-distributed Stochastic Neighbor Embedding (t-SNE)~ \cite{maaten2008visualizing}. We utilize color coding to annotate samples from different classes. 

\begin{figure}[t]
\centering
    {\includegraphics[width=6.5cm]{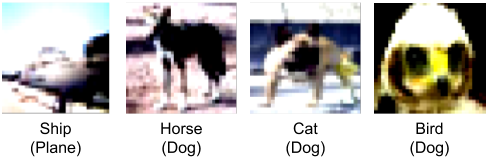} }%
\caption {Sample images identified by our framework that are potentially mislabeled in CIFAR10 testset. Text below each image denotes the ground-truth label provided by CIFAR10 and text in parenthesis are the predicted labels by our framework.}
\label{img:mislabelTest}
\vspace{-0.6cm}
\end{figure}

We visualize how data distribution is affected by noise in Fig.~\ref{img:tsne}, where Fig.~\ref{img:tsne}(a) presents the visualization of CIFAR10 training set features without any artificial noise, yet we can observe a very small amount of noise. We hypothesise that similar noise can be present across different annotated datasets, targeting a variety of tasks, available today. When we introduce 20\% random label noise to the dataset, the samples get more scattered across the feature space (Fig.~\ref{img:tsne}(b)). Our framework can identify these noisy samples and effectively clean them as demonstrated in Fig.~\ref{img:tsne}(c). We also present the visualization of test samples from CIFAR10 in Figs.~\ref{img:tsne}(d)-(f). Training with noisy data adversely affects the DNN's ability to extract features robustly (Fig.~\ref{img:tsne}(d)). If data denoising is performed prior to training a DNN, we can minimize this adverse effect greatly (Fig.~\ref{img:tsne}(e)). However, data distributions still cannot be very concise and often overlap. This phenomenon can be explained as even if there are no noise in a training set and the test set might still contain noise and confusing samples. Our framework can filter out most of the boundary samples from respective distributions, as demonstrated in Fig.~\ref{img:tsne}(f). Yet, if we closely observe, our framework missed some samples what are well within the distribution but predicted labels do not match the ground-truth labels. We argue that these samples share dominant features with samples from the closest distribution or may be mislabeled, as we can similarly observe in clean training data distribution (Fig.~\ref{img:tsne}(a)). We have presented some examples of potentially mislabeled test samples of CIFAR10 in Fig.~\ref{img:mislabelTest}.

\begin{figure}[t]
\centering
    {\includegraphics[width=9cm]{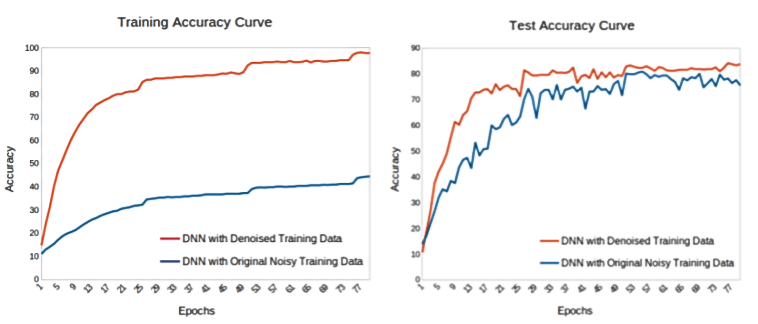} }
    \caption {Effect of our proposed framework on training ResNet34 with the original noisy data (60\% label noise) and denoised data. (left) Learning curves on CIFAR10 training data; (right) Learning curves on CIFAR10 test data.}
\label{img:denoise}
\vspace{-0.5cm}
\end{figure}

Fig.~\ref{img:denoise} provides further evidence of the effectiveness of our proposed framework. In this plot we show how denoising helps accelerate learning of DNNs. In this experiment, we train ResNet34 on CIFAR10 introducing heavy label noise (60\%). Fig.~\ref{img:denoise} shows a stark difference between learning from original noisy data vs. denoised data by our framework. Our framework not only accelerates learning (left), but also improves accuracy on test data when learning from denoised data (right). Evidently, our framework can effectively clean data and expedite learning by eliminating noisy or confusing samples.

\section{Conclusion}\label{conclusion}

Noisy data is one of the most crucial hurdles for DNNs to achieve high accuracy and reliable performance. In this paper, with rigorous experimentation, we have shown that complicated, specialized training to filter out noise in data is not always effective and necessary. On the contrary, we show features learned by off-the-shelf DNNs are quite robust. With a simple yet effective filtering mechanism, we can achieve competitive, often better, performance than these specialized models. However, we would like to point out some limitations and future work of our proposed framework. We consider threshold based on distance from distributions as a filtering criteria. While it has proven to be very successful, a distance threshold will limit data distribution to be spherical, but in reality data distributions can often be irregular. This can explain why our framework sometimes does not perform as expected. A more robust filtering method requires a more accurate model of distribution. One other pathway to address this issue would be learning more robust features along with filtering techniques. We leave these areas open for future research.

\bibliographystyle{splncs}
\bibliography{egbib}

\end{document}